\documentclass[letterpaper]{article} 
\usepackage{aaai2026}  
\usepackage{times}  
\usepackage{helvet}  
\usepackage{courier}  
\usepackage[hyphens]{url}  
\usepackage{graphicx} 
\usepackage{natbib}  
\usepackage{caption} 
\usepackage{algorithm}
\usepackage{algorithmic}

\usepackage{newfloat}
\usepackage{listings}
\DeclareCaptionStyle{ruled}{labelfont=normalfont,labelsep=colon,strut=off} 
\floatstyle{ruled}
\newfloat{listing}{tb}{lst}{}
\floatname{listing}{Listing}
\usepackage{amsmath}
\usepackage{amssymb}    
\usepackage{booktabs}
\usepackage{multirow}
\usepackage[table,xcdraw]{xcolor}
\usepackage{tabularx}
\usepackage{algorithm}
\usepackage{algorithmic}

\usepackage[breakable]{tcolorbox}  
\usepackage{graphicx}
\usepackage{xcolor}
\definecolor{my_green}{RGB}{0,128,0}
\definecolor{my_blue}{RGB}{0,0,255}
\definecolor{my_purple}{RGB}{128,0,128}

\title{RPM-MCTS: Knowledge-Retrieval as Process Reward Model \\ with Monte Carlo Tree Search for Code Generation}

\author {
    Yuanyuan Lin\textsuperscript{\rm 1,\rm 3}\equalcontrib,
    Xiangyu Ouyang\textsuperscript{\rm 2}\equalcontrib,
    Teng Zhang\textsuperscript{\rm 1}\thanks{Corresponding Author},
    Kaixin Sui\textsuperscript{\rm 3}
}
\affiliations {
    \textsuperscript{\rm 1} School of Computer Science and Technology, Huazhong University of Science and Technology, China \\
    \textsuperscript{\rm 2} College of Computer Science and Technology, Xi'an Jiaotong University, China \\
    \textsuperscript{\rm 3} ByteDance Seed, China \\
    \{linyy, tengzhang\}@hust.edu.cn, oyxy2019@stu.xjtu.edu.cn, suikaixin@163.com
}

\begin{document}
\maketitle

\begin{abstract}
    Tree search-based methods have made significant progress in enhancing the code generation capabilities of large language models. However, due to the difficulty in effectively evaluating intermediate algorithmic steps and the inability to locate and timely correct erroneous steps, these methods often generate incorrect code and incur increased computational costs. To tackle these problems, we propose RPM-MCTS, an effective method that utilizes Knowledge-\underline{R}etrieval as \underline{P}rocess Reward \underline{M}odel based on \underline{M}onte \underline{C}arlo \underline{T}ree \underline{S}earch to evaluate intermediate algorithmic steps. By utilizing knowledge base retrieval, RPM-MCTS avoids the complex training of process reward models. During the expansion phase, similarity filtering is employed to remove redundant nodes, ensuring diversity in reasoning paths. Furthermore, our method utilizes sandbox execution feedback to locate erroneous algorithmic steps during generation, enabling timely and targeted corrections. Extensive experiments on four public code generation benchmarks demonstrate that RPM-MCTS outperforms current state-of-the-art methods while achieving an approximately 15\% reduction in token consumption. Furthermore, full fine-tuning of the base model using the data constructed by RPM-MCTS significantly enhances its code capabilities.
\end{abstract}


\section{Introduction}

Code generation aims at understanding problem descriptions in natural language and generating the corresponding code snippets. In recent years, large language models (LLMs) have demonstrated remarkable performance in code generation tasks~\cite{zhang2024unifying}. For code generation, the early methods involve dividing code planning and synthesis into two phases using chain-of-thought or tree structures~\cite{wei2022chain, jiang2024self, zelikman2023parsel}. \citet{wang2024planning} have demonstrated that providing LLMs with \emph{a correct} solution can significantly enhance model performance, even when these solutions consist of incomplete plans, i.e., for solutions, correctness is preferred over completeness, and the key to enhancing the code generation capability of LLMs lies in generating correct plans.

Programming languages possess their own inherent logical structures and tightly interconnected knowledge, which makes it essential not to overlook long-range dependencies within the code. Previous work has shown that rotary position embedding does not always lead to attention weights decaying with relative distance~\cite{barbero2024round}. Concurrently, through exploring the attention distribution between tokens, we have experimentally demonstrated that selecting algorithmic steps as the basic units is a superior choice. Therefore, our objective focuses on how to accurately generate intermediate algorithmic steps.

However, a limitation of previous methods lies in the lack of an evaluation and correction mechanism for intermediate algorithmic steps, which fails to guarantee the correctness of these steps~\cite{lu2025lsr,li2025structured}. One way to tackle this issue is to use a value function or reward model to verify reasoning traces for correctness, which then serves as a learning signal for self-training~\cite{lightman2023let,wang2023math}. However, training a reliable reward model to verify every step in a reasoning trace generally depends on dense human-generated annotations per reasoning step~\cite{lightman2023let}, which does not scale well.

Unlike other reasoning tasks, code generation benefits from the homogeneity of algorithmic workflows across different problem categories. This allows us to leverage historical experience from a knowledge base containing numerous correct algorithmic steps to evaluate the process reward of expansion steps. Additionally, code generation typically benefits from detailed feedback provided by compilers. Consequently, in this paper, we propose RPM-MCTS, which optimizes the Monte Carlo Tree Search (MCTS) algorithm using external information feedback. Our method utilizes the knowledge base for intermediate algorithmic step-level evaluation and employs sandbox feedback for result-level assessment of complete code. Specifically, the root node of the Monte Carlo tree represents the coding problem, while all other nodes represent individual algorithmic steps. During each iteration, multiple distinct potential next steps are generated based on the current reasoning path. Node selection is guided by historical experience from the knowledge base, enabling faster discovery of high-value search paths. In the simulation phase, complete code is generated and evaluated using sandbox and model feedback to update node values. Notably, during simulation, we localize erroneous steps within the full algorithmic workflow and incorporate newly generated correct steps into the tree, thereby reducing token consumption. After multiple iterations, the highest-scoring path from root to leaf is selected, ultimately yielding a complete solution alongside its corresponding code. The contributions are summarized as follows:
\begin{itemize}
    \item We propose RPM-MCTS, which leverages knowledge base retrieval scores to evaluate intermediate algorithmic steps, steering LLMs to explore high-value reasoning paths more effectively.
    \item We leverage sandbox feedback during the simulation phase to evaluate code generated from reasoning steps, localize errors, and truncate simulations, thereby reducing computational costs.
    \item We conduct extensive experiments and show that RPM-MCTS is superior to state-of-the-art methods. Moreover, we verify that base models fine-tuned with data generated by RPM-MCTS enjoy greater code capabilities.
\end{itemize}

\section{Related Work}

\subsubsection{Monte Carlo Tree Search.}

As the extension of Chain-of-Thought (CoT)~\cite{wei2022chain}, Tree-of-Thought (ToT)~\cite{yao2023tree} enhances the reasoning and planning capabilities of LLMs by exploring different thought paths within a tree structure. Subsequently, Monte Carlo Tree Search has served as a search algorithm to more effectively guide LLMs in exploring intermediate sub-steps~\cite{zhao2023large,hao2023reasoning,zhou2023language,ding2023everything}. ReST-MCTS*~\cite{zhang2024rest} combines process reward guidance with Monte Carlo Tree Search to collect high-quality reasoning trajectories and step-by-step values for training strategy and reward models. SRA-MCTS~\cite{xu2024sra} further extends this to the field of code generation, using Monte Carlo Tree Search to generate intermediate reasoning steps and conducting iterative self-evaluation to synthesize training data for supervised fine-tuning. However, relying solely on model self-evaluation introduces biases and hallucinations, and small-scale LLMs exhibit limited instruction-following capabilities. RethinkMCTS~\cite{li2025rethinkmcts} is another prior work that also uses execution feedback but employs a patching strategy. If this patch fails, the search may proceed on an incorrect path, making it less suitable for generating high-quality SFT data.

\subsubsection{Process Evaluation.} 

In heuristic search, a robust reasoning process needs to have self-evaluation capabilities, and the evaluation results are further used to guide the search. Early work mainly focused on outcome-level evaluation~\cite{cobbe2021training}, that is, evaluating the complete solution after the reasoning is completed. Outcome-level evaluation is simple to implement but often requires more detailed assessment. Step-level evaluation~\cite{lightman2023let,wang2023math,gao2024llm} emphasizes the assessment of individual reasoning steps. In tree search algorithms, process evaluation is widely used to guide search trajectories. Logic-RL~\cite{xie2025logic} optimizes path selection by implementing state scoring in beam search. Furthermore, step-level evaluation has proven its effectiveness in both error correction and the summarization of reasoning steps. \citet{zheng2024makes} developed a method capable of accurately locating inaccuracies in specific reasoning steps, thereby providing more precise and actionable feedback for comprehensive evaluation.

\section{Method}

In this section, we elaborate on the proposed modified MCTS that incorporates the knowledge base as a process reward model. The methodology comprises three key components: knowledge base construction, RPM-MCTS, and code generation. First, knowledge base retrieval scores circumvent random selection during node expansion. Then, in the expansion phase, nodes are filtered based on similarity metrics to eliminate redundant candidates. Finally, during the simulation phase, the algorithm performs error reflection and retains nodes with verified correct reasoning. These collective strategies enable faster exploration of higher-quality algorithmic steps.

\subsection{Knowledge Base Construction}

In this section, we introduce the construction of a retrievable global knowledge base designed to mitigate hallucination during the planning process. Due to the homogeneity of algorithms within the same category, where fundamental principles and methods are relatively similar, we utilize a knowledge base containing numerous correct algorithms across diverse categories. This serves as the evaluation model for intermediate algorithmic steps in RPM-MCTS, eliminating the need to train a separate process reward model.

We use the training set data from APPS~\cite{hendrycks2021measuring} and CodeContests~\cite{li2022competition}, which contain coding problems paired with their correctly implemented solutions.  We utilize the Claude Sonnet 3.7 to generate the correct algorithmic steps corresponding to the correct code and decompose them step by step. We sequentially concatenate the problems by rolling them out according to the algorithmic steps. Specifically, for problem $p_i$ with $n_i$ algorithmic steps and $a_{i}^{(j)}$ corresponding to the $j$-th step, we have
\begin{align}
    \mathcal{K}_i = \{ \mathrm{concat}(p_i, a_i^{(1)}, \ldots, a_i^{(j)}), ~ j = 1, 2, \ldots, n_i \},
\end{align}
and $\mathcal{K} = \uplus_{i = 1}^n \mathcal{K}_i$ is the knowledge base with cardinality $n$.

To enhance retrieval efficiency and improve retrieval precision by distinguishing between problems with similar descriptions but different algorithmic solutions, we organize the knowledge base into 14 distinct algorithm categories and store them as vector database by using the BGE~\cite{xiao2024c} embedding model.

\subsection{RPM-MCTS}

We propose an enhanced MCTS method, named RPM-MCTS. In this method, the root node represents the problem, while all other nodes represent an algorithmic step. Specifically, the method comprises four distinct phases: Selection, Expansion, Evaluation and Reflection, and Backpropagation, as shown in Figure~\ref{fig1}. These phases are performed on a search tree composed of tree nodes and are iterated multiple times, with each iteration generating a concrete algorithmic step.

\begin{figure*}[ht]
    \centering
    \includegraphics[width=0.8\textwidth]{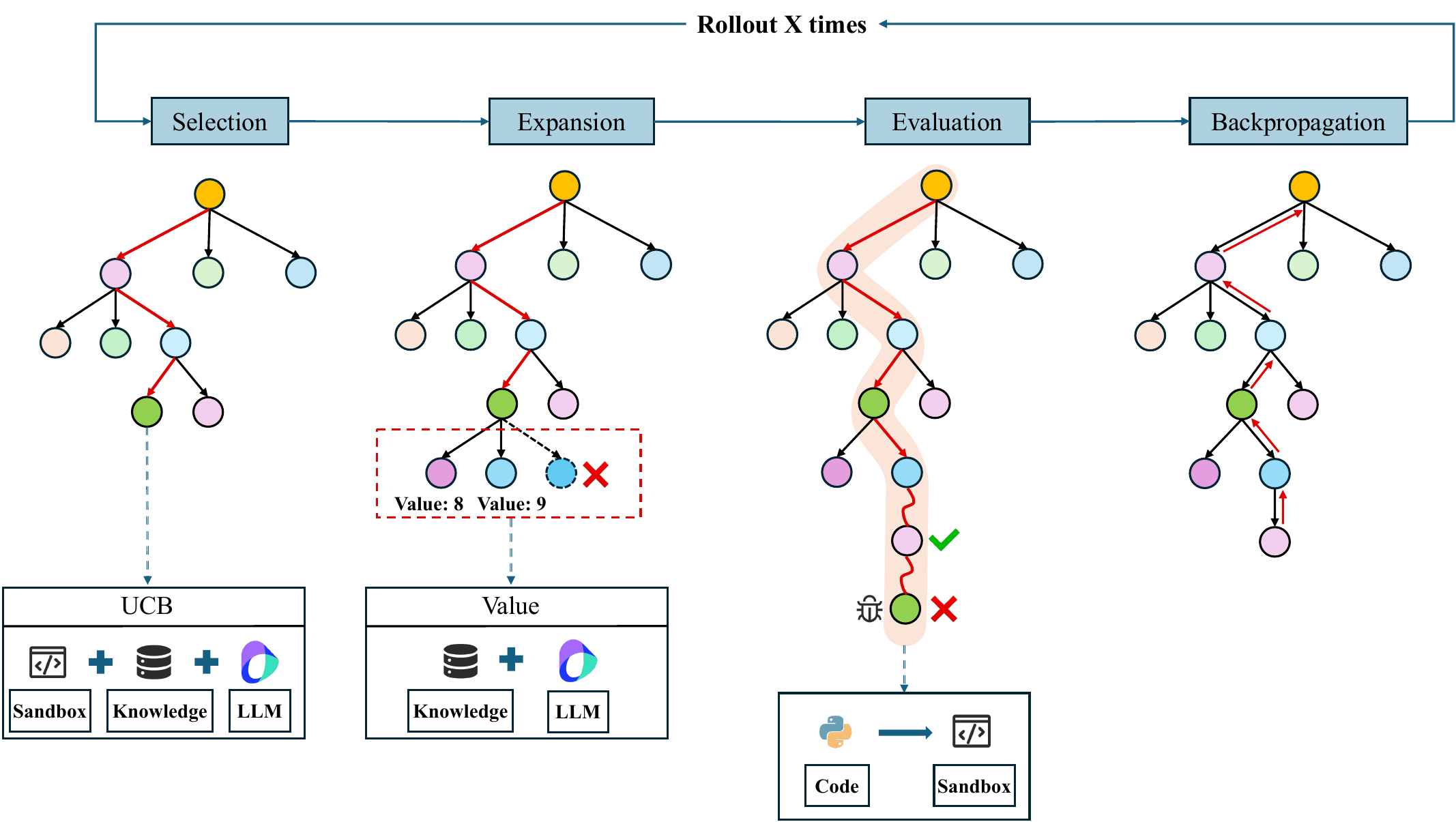} 
    \caption{Overview of RPM-MCTS. (a) Selection: Select a leaf node according to Eqn.~\eqref{eq-action-total-value}. (b) Expansion: After selecting a node, expand multiple child nodes, and use knowledge base retrieval scores and LLM evaluation to select nodes for simulation. The node color represents similarity magnitude. (c) Evaluation: Generate complete reasoning steps for the selected node, generate code strictly in accordance with these reasoning steps, and use a sandbox for information feedback. (d) Backpropagation: Propagate the reward scores backward. The yellow root node represents the problem, and the remaining nodes represent each reasoning step.}
    \label{fig1}
\end{figure*}

\subsubsection{Selection.} 

In the selection phase, a leaf node is selected from the current tree for further expansion according to the selection score, which is defined as a weighted combination of the Upper Confidence Bound (UCB)~\cite{silver2017mastering} and the knowledge base retrieval score:
\begin{align} \label{eq-action-total-value}
    \mathrm{SelectionScore}(s,a) = \mathrm{UCB}(s,a) + \alpha K(s,a),
\end{align}
where $(s,a)$ denotes a state-action pair with $s$ containing the description of the problem and previously generated algorithmic steps and $a$ representing the new step at the current node. The parameter $\alpha$ is for balancing the two terms. 

UCB is a classical multi-armed bandit algorithm and well performed in addressing the exploration-exploitation trade-off. UCB selects actions by computing an upper confidence estimate of each action’s potential reward: 
\begin{align}
    \mathrm{UCB}(s,a) = Q(s,a) + \beta \sqrt{\frac{\log N(s)}{1+N(s,a)}},
\end{align}
where $Q(s,a)$ represents the empirical mean cumulative reward after taking action $a$ from state $s$, $N(s)$ is the number of times state $s$ has been explored in the current context, and $N(s,a)$ is the number of times action $a$ has been taken in state $s$. The parameter $\beta$ is for trading off the exploitation (the former term) and exploration (the latter term).

The knowledge base retrieval score $K(s,a)$ is obtained by retrieving the concatenated $(s,a)$ pair from the knowledge base. Specifically, let $f$ denote the embedding model that maps $(s,a)$ to a vector with the same dimension as the knowledge base. Given the preceding reasoning path, the knowledge base retrieval score for the current node is calculated as follows:
\begin{align}
    K(s,a) = \max \left( 0, \max_{k \in \mathcal{K}} \frac{f((s,a)) \cdot k}{ \| f((s,a)) \| \cdot \| k \|} \right).
\end{align}
The knowledge base similarity score $K(s,a)$ enables acquisition of step-wise assessments prior to the evaluation phase. In other words, when newly generated nodes remain unexplored, we prioritize leveraging historically validated solutions through knowledge base retrieval scores to identify higher-value nodes.

Starting from the root node, we recursively select the child node with the maximum $\mathrm{SelectionScore}$ value at each branching point. Selection ties are resolved stochastically. Each iteration advances to the highest-scoring child node until reaching a leaf node.

\begin{figure}[ht]
    \centering
    \includegraphics[width=1.0\columnwidth]{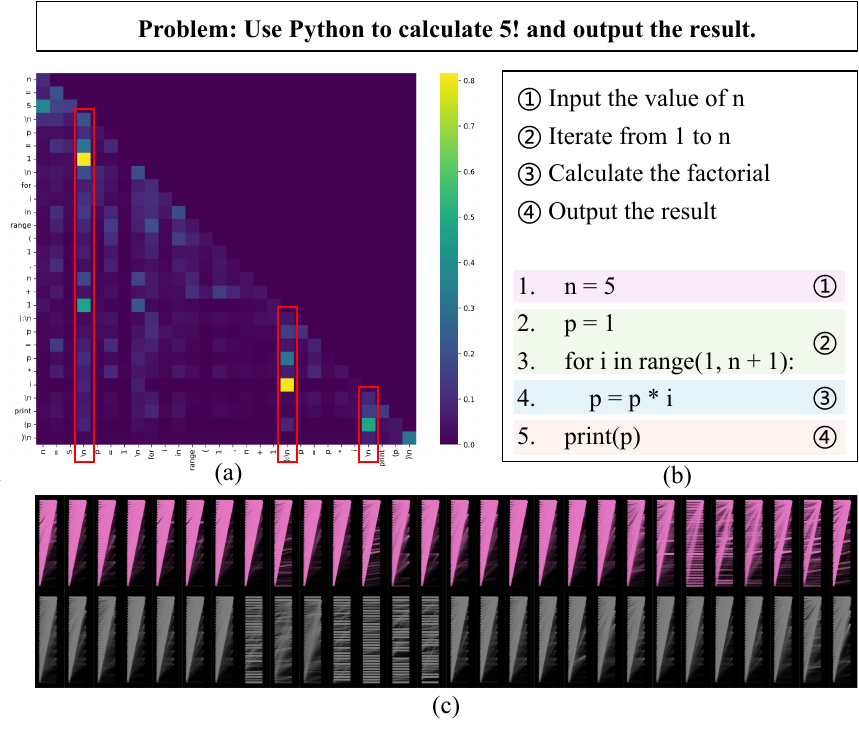} 
    \caption{(a) Token-level attention heatmap for code corresponding to the programming problem. (b) Algorithmic steps and corresponding code for the programming problem. (c) Attention sink phenomenon.}
    \label{fig2}
\end{figure}

\subsubsection{Expansion.}

Upon selecting a leaf node during the selection phase, the expansion phase aims to generate remaining child nodes, thereby expanding the search scope of the entire tree. Since the attention weights between tokens do not always decay with relative distance~\cite{barbero2024round}, we conduct an in-depth study on the attention mechanism between tokens in LLMs during code generation to reveal the influencing factors among tokens. As shown in Figure~\ref{fig2}, it can be observed that certain tokens have a profound impact on subsequent code generation. It can thus be inferred that these key tokens can summarize and interpret the information of previously generated tokens, and have higher reference value for subsequent token generation. Meanwhile, relevant studies~\cite{barbero2025llms,xiao2023efficient} have shown that modern LLMs exhibit the phenomenon of ``attention sink''. Specifically, numerous attention heads allocate a disproportionate share of weights, such as exceeding 30\% or even 80\%, to the beginning-of-sequence token ⟨bos⟩, despite its primary function as a sequence delimiter with minimal semantic content. Therefore, to facilitate our examination of inter-token dependencies in code generation tasks, we selectively visualize token attention mechanisms at designated layers. Figure~\ref{fig2}(c) shows that attention not only sinks to ⟨bos⟩ but also peaks at algorithmic step boundaries, justifying that algorithmic step blocks are more effective basic processing units in code generation tasks. Therefore, we select algorithmic steps as the basic units for expansion.

To ensure diversity in generated steps during the expansion phase, we implement a sampling decoding strategy that sequentially generates each child node. Specifically, to prevent repetitive generation by the LLM, we iteratively provide all previously generated steps as context when producing each new step. The input for the LLM is
\begin{align}
    \mathrm{concat} (s, a_1, \ldots, a_i, g), ~ i = 1, 2, \ldots, b
\end{align}
where $g$ represents the reflection in the simulation phase, and $b$ denotes the maximum number of branches each node can expand.

After expanding $b$ nodes, we employ cosine similarity for filtering to reduce computational costs by avoiding simulations on redundant nodes. Specifically, we map the reasoning steps of the $b$ nodes to vectors using the embedding model $\mathcal{E}$ and calculate the cosine similarities between these $b$ nodes. When the similarity exceeds a predetermined threshold, the node is identified as redundant and filtered out. This method effectively reduces the search space and enhances algorithmic efficiency while maintaining diversity.

\subsubsection{Evaluation and Reflection.}

During the evaluation phase, simulation and evaluation are performed for the selected leaf nodes. We provide the LLM with the algorithmic steps $s$ already generated for the node and its ancestor nodes, enabling the LLM to strictly follow the generated steps and continue simulating to complete all remaining steps. We search for the thoughts and evaluate with the code generated following the thoughts.

The generated code undergoes sandbox evaluation using public test cases. However, since public test cases only cover a subset of possible scenarios, the code may fail on unseen cases, such as boundary conditions or performance issues. We therefore employ the LLM to analyze the complete algorithmic steps based on sandbox feedback.

We assess the steps generated during the expansion phase through two components, which are the pass rate on public test cases and LLM evaluation. The final evaluation score is obtained by weighted summation of these two scores. The formula is as follows:
\begin{align}
    Q(s,a)=\gamma \cdot r_{\text{exec}} + (1- \gamma) \cdot r_{\text{LLM}}
\end{align}
where $r_{\text{exec}}$ denotes the pass rate on public test cases, $r_{\text{LLM}}$ represents the score from LLM evaluation based on the sandbox feedback results and complete steps provided to the LLM, and $\gamma$ indicates the weight controlling these two parts of the scores.

For code that fails to public test cases, we isolate erroneous algorithmic steps by decomposing the code into blocks and sequentially debugging each block via LLM analysis with public test inputs~\cite{zhong2024debug}. We retain all correct steps generated during the simulation phase, truncated before the first erroneous step. These validated steps are then incorporated into the MCTS tree as expanded nodes.

The entire RPM-MCTS process is terminated when the solution passes all public test cases and achieves a high LLM evaluation score. Otherwise, node updates and reflection are performed, and the RPM-MCTS process proceeds until the maximum iteration count is reached.

\subsubsection{Backpropagation.}

The objective of backpropagation is to update the reward values of nodes upon completion of state value evaluation. We propagate reward values backward from leaf nodes to the root node, updating the state estimates of all nodes along the path. For newly generated nodes during the expansion phase, they collectively update their parent node. As the number of simulations increases, these value estimates become increasingly accurate. This process repeats until the preset maximum simulation count is reached, ultimately resulting in a search tree that records the state value and visit count for each node.

\subsection{Generate Code}

Termination of the RPM-MCTS process occurs under two conditions: 1) If all public test cases are passed and LLM analysis confirms robustness to unseen edge cases before reaching maximum iterations, the code generated during the simulation phase is retained. 2) When maximum iterations are reached without meeting termination criteria, the leaf node with the highest state value is selected, its ancestral path is traced, and the LLM is instructed to generate code by rigorously adhering to the algorithmic steps assembled from this path.

\section{Experiments}

\subsection{Experimental Settings}

\subsubsection{Datasets.}

For the construction of the knowledge base, we use the train set splits of APPS~\cite{hendrycks2021measuring} and CodeContests~\cite{li2022competition} as data sources. After validation and filtering, we obtained 11,038 samples with a total of 82,923 steps. For benchmarking, we used the test set splits of APPS and CodeContests, as well as HumanEval+~\cite{liu2023your} and MBPP+~\cite{liu2023your}. The APPS dataset contains three difficulty levels: introductory, interview, and competition. We selected 150 validated samples from each difficulty level. The CodeContests dataset consists of competitive programming problems collected from contest websites such as Codeforces. Additionally, HumanEval~\cite{chen2021evaluating} and MBPP~\cite{austin2021program} are widely recognized benchmarks in the code generation domain, while HumanEval+ and MBPP+ introduce a larger number of test cases to enable more accurate evaluations. We utilized Claude Sonnet 3.7 to convert all datasets into a unified format, which primarily includes the problem statement, public test cases, private test cases, and standard solution. To facilitate sandbox execution, we transformed datasets with standard input-output problems into function definitions with docstrings. For datasets without public test cases, we selected the first two private test cases as public test cases.

\subsubsection{Baselines.}

We selected the following methods as baselines for comparison. Base LLM refers to directly prompting the LLM to output solution code using the problem statement and public test cases as input. LDB~\cite{zhong2024debug} leverages the LLM to track intermediate variables during code execution to iteratively improve the code. ToT~\cite{yao2023tree} performs a search of thought steps using DFS or BFS before generating the final code. SRA-MCTS~\cite{xu2024sra} combines LLM with MCTS to explore intermediate reasoning steps. The complete steps obtained by SRA-MCTS are used as input to prompt the LLM to directly infer and output the solution code for evaluation.

\subsubsection{Implementation Details.}

We use two large-parameter backbone models, Qwen3-235B-A22B~\cite{yang2025qwen3} and Claude Sonnet 3.7, alongside a smaller-parameter model, Qwen3-8B. In the code generation domain, pass@k~\cite{chen2024survey} is a widely used metric, and we adopted pass@1 as the evaluation metric. The rollout, i.e., maximum number of iterations, was set to 5 for all methods. The branching factor $b$ for tree-based methods was set to 3. The exploration constant $\beta$ for UCB was set to 0.5. In RPM-MCTS, the weight of the knowledge base retrieval score $\alpha$ was set to 0.5, and the similarity filtering threshold was set to 0.85.

\subsection{Main Results}

\begin{table*}[t]
    \centering
    {
        \small  
        \begin{tabular}{lrrrrrrr}
            \toprule 
            \multicolumn{1}{c}{\textbf{Method}} & \multicolumn{1}{c}{\textbf{APPS-Intro.}} & \multicolumn{1}{c}{\textbf{APPS-Interv.}} & \multicolumn{1}{c}{\textbf{APPS-Comp.}} & \multicolumn{1}{c}{\textbf{CodeContests}} & \multicolumn{1}{c}{\textbf{HumanEval+}} & \multicolumn{1}{c}{\textbf{MBPP+}} & \multicolumn{1}{c}{\textbf{Average}} \\
            \midrule 
            \multicolumn{8}{c}{\textbf{Qwen3-8B}}                                                                                                                                                                                                                                                                                                  \\ 
            \midrule
            Base LLM                            & 56.7                                     & 35.3                                      & 29.3                                    & 10.7                                      & 75.6                                    & 72.2                               & 52.1                                 \\
            LDB                                 & 64.0 (+7.3)                              & 42.0 (+6.7)                               & 28.0 (-1.3)                             & 11.3 (+0.7)                               & 78.1 (+2.4)                             & 70.1 (-2.1)                        & 53.5 (+1.4)                          \\
            ToT                                 & 69.3 (+12.7)                             & 54.0 (+18.7)                              & 41.3 (+12.0)                            & 17.3 (+6.7)                               & 82.3 (+6.7)                             & 70.4 (-1.9)                        & 59.0 (+6.9)                          \\
            SRA-MCTS                            & 67.3 (+10.7)                             & 42.7 (+7.3)                               & 29.3 (+0.0)                             & 16.0 (+5.3)                               & 73.8 (-1.8)                             & 65.9 (-6.4)                        & 52.8 (+0.7)                          \\
            Ours w/o KB                         & 76.7 (+20.0)                             & 56.7 (+21.3)                              & 40.7 (+11.3)                            & 22.3 (+11.6)                              & 82.3 (+6.7)                             & \textbf{78.3 (+6.1)}               & 63.5 (+11.4)                         \\
            Ours                                & \textbf{77.3 (+20.7)}                    & \textbf{60.0 (+24.7)}                     & \textbf{43.6 (+14.3)}                   & \textbf{23.0 (+12.3)}                     & \textbf{83.5 (+7.9)}                    & 76.2 (+4.0)                        & \textbf{64.0 (+11.9)}                \\
            \midrule
            \multicolumn{8}{c}{\textbf{Qwen3-235B-A22B}}                                                                                                                                                                                                                                                                                           \\ 
            \midrule
            Base LLM                            & 78.0                                     & 54.7                                      & 42.7                                    & 24.0                                      & 86.0                                    & 78.8                               & 64.6                                 \\
            LDB                                 & 78.7 (+0.7)                              & 61.3 (+6.7)                               & 47.3 (+4.7)                             & 25.3 (+1.3)                               & 86.0 (+0.0)                             & 78.8 (+0.0)                        & 66.4 (+1.8)                          \\
            ToT                                 & 84.7 (+6.7)                              & 62.7 (+8.0)                               & 57.3 (+14.7)                            & 27.3 (+3.3)                               & 85.4 (-0.6)                             & 75.4 (-3.4)                        & 67.7 (+3.1)                          \\
            SRA-MCTS                            & 76.0 (-2.0)                              & 52.7 (-2.0)                               & 44.0 (+1.3)                             & 24.7 (+0.7)                               & 85.4 (-0.6)                             & 70.9 (-7.9)                        & 61.7 (-3.0)                          \\
            Ours w/o KB                         & \textbf{88.0 (+10.0)}                    & \textbf{72.0 (+17.3)}                     & 52.0 (+9.3)                             & 34.7 (+10.7)                              & 86.6 (+0.6)                             & 79.9 (+1.1)                        & 71.3 (+6.7)                          \\
            Ours                                & 86.7 (+8.7)                              & 67.3 (+12.7)                              & \textbf{59.3 (+16.7)}                   & \textbf{36.7 (+12.7)}                     & \textbf{87.8 (+1.8)}                    & \textbf{81.2 (+2.4)}               & \textbf{72.3 (+7.7)}                 \\
            \midrule
            \multicolumn{8}{c}{\textbf{Claude Sonnet 3.7}}                                                                                                                                                                                                                                                                                         \\ 
            \midrule
            Base LLM                            & 78.7                                     & 56.0                                      & 59.3                                    & 31.3                                      & 82.9                                    & 77.8                               & 67.3                                 \\
            LDB                                 & 82.0 (+3.3)                              & 64.7 (+8.7)                               & 73.3 (+14.0)                            & 33.3 (+2.0)                               & 88.4 (+5.5)                             & 77.0 (-0.8)                        & 71.5 (+4.2)                          \\
            ToT                                 & 84.0 (+5.3)                              & 68.0 (+12.0)                              & 66.0 (+6.7)                             & 39.3 (+8.0)                               & 86.0 (+3.1)                             & 74.6 (-3.2)                        & 70.8 (+3.6)                          \\
            SRA-MCTS                            & 83.3 (+4.7)                              & 63.3 (+7.3)                               & 62.0 (+2.7)                             & 36.0 (+4.7)                               & 81.1 (-1.8)                             & 74.3 (-3.4)                        & 68.4 (+1.1)                          \\
            Ours w/o KB                         & \textbf{92.0 (+13.3)}                    & 73.3 (+17.3)                              & 78.0 (+18.7)                            & 42.7 (+11.3)                              & 86.6 (+3.7)                             & 79.1 (+1.3)                        & 76.2 (+8.9)                          \\
            Ours                                & \textbf{92.0 (+13.3)}                    & \textbf{74.0 (+18.0)}                     & \textbf{81.3 (+22.0)}                   & \textbf{46.0 (+14.7)}                     & \textbf{89.0 (+6.1)}                    & \textbf{81.0 (+3.2)}               & \textbf{78.1 (+10.9)}                \\
            \bottomrule 
        \end{tabular}
    } 
    \caption{Performance comparison of all methods across different backbone models on code generation benchmarks. Values in parentheses indicate the improvement over the base LLM.}
    \label{table1}
\end{table*}

Our method achieves the most significant improvements across different backbone models and datasets. As shown in Table \ref{table1}, Qwen3-8B achieves an average improvement of 11.90\%, Qwen3-235B-A22B achieves an average improvement of 7.71\%, and Claude Sonnet 3.7 achieves an average improvement of 10.86\%. Since the base Qwen3-8B performs worse than the other two larger base LLMs across all datasets, especially on simpler datasets, the Qwen3-8B shows the most significant improvement when using RPM-MCTS. On the two more challenging datasets, APPS-competition and CodeContests, Qwen3-8B achieves an average improvement of 13.3\%, Qwen3-235B-A22B achieves an average improvement of 14.67\%, and Claude Sonnet 3.7 achieves an average improvement of 18.34\%. This is because Qwen3-8B has weaker evaluation scoring capabilities, while larger LLMs have relatively stronger evaluation capabilities, resulting in greater gains. This demonstrates that the more difficult the task, the more accurate evaluation of intermediate algorithm steps is required.

LDB achieves greater improvements on simpler datasets compared to more challenging ones. We found that this is because, for more difficult problems, through multiple rounds of execution feedback, LLMs often only modify code conditions to pass public test cases rather than thinking about modifying the actual logic of the code. SRA-MCTS shows performance improvements on more challenging datasets but declines on simpler ones. The reason is that for simple problems, LLM evaluation scores are always perfect or near-perfect, prematurely ending the search for steps, resulting in incomplete or lower-quality reasoning steps.

Comparing the results across three different difficulty levels in the APPS dataset, it can be observed that for the two larger LLMs, as the difficulty increases, our method brings more significant performance improvements. The higher the difficulty of the problem, the more guidance the LLM needs to avoid getting lost in complex reasoning chains. This demonstrates the effectiveness of our method in evaluating intermediate steps, helping LLMs enhance their evaluation capabilities and further unlocking the vast potential code knowledge and reasoning abilities inherent in LLMs.

For fair comparison, even without using knowledge base retrieval scores as rewards, our method outperforms other baselines. Experimental results show that overall, especially on the two most challenging datasets, incorporating the knowledge base further stabilizes and improves performance. The reason is that LLM evaluation of intermediate steps in complex problems is unreliable, and random exploration struggles to find the correct solution path. Therefore, leveraging the knowledge base to use the reasoning patterns of historically similar problems as guidance helps direct the search. This demonstrates the effectiveness of using knowledge base retrieval scores as rewards for intermediate process evaluation.

On a few simpler datasets, performance slightly improves when knowledge base retrieval scores are not used. We analyze that this is because, in simple tasks, LLMs can already accurately evaluate the quality of generated paths. In this case, introducing knowledge base rewards, while aiming to provide additional prior information, may retrieve historical cases that are textually similar but logically different in their solutions, introducing noise into MCTS node selection. In contrast, for complex tasks, LLM evaluation capabilities for intermediate steps are limited, the search space is vast, and solutions are sparse. The structured priors provided by the knowledge base effectively guide the search direction, significantly improving success rates. This phenomenon indicates that the effectiveness of knowledge base rewards depends on the balance between task difficulty and LLM evaluation confidence.

\subsection{Ablation Study}

We conduct ablation experiments using Qwen3-235B-A22B as the backbone model to evaluate performance, and the results are shown in Figure \ref{fig3}.

\textbf{w/o KB} indicates that only LLM evaluation is used in selection, without knowledge base retrieval. Compared to the complete method, the overall performance slightly decreased, with an average drop of 1.05\%. The decline was most significant on the two more challenging datasets, with an average drop of 4.67\%. This indicates that large models still face challenges with complex problems. By introducing a knowledge base to compare the generated reasoning steps with the correct reasoning steps of similar problems in the knowledge base, the self-assessment capability for complex problems can be improved.

\textbf{w/o ER} means that the execution rewards of public test cases in the sandbox are not used during the simulation phase. This resulted in the largest overall performance drop, highlighting that the core of RPM-MCTS reflection lies in the detailed feedback provided by the code execution environment. In fact, previous research~\cite{huang2023large} has already pointed out that without external feedback, LLMs lack the ability to self-correct their reasoning processes.

\begin{figure}[h]
    \centering
    \includegraphics[width=1.0\columnwidth]{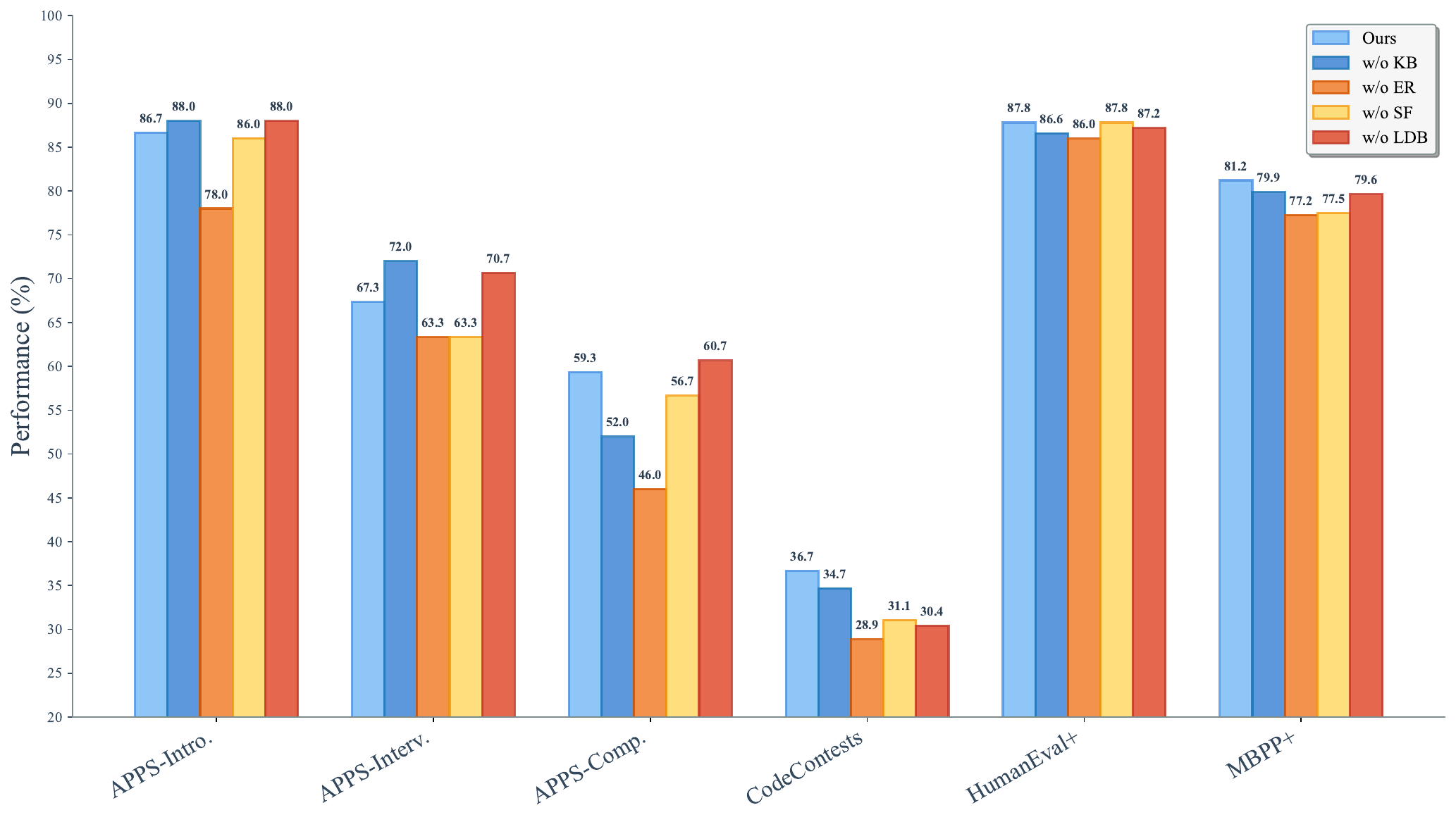} 
    \caption{Ablation study results on different benchmarks.}
    \label{fig3}
\end{figure}

\textbf{w/o SF} refers to the removal of similarity filtering, i.e., not discarding similar child nodes during the expansion phase. The results show that filtering out repeated intermediate algorithmic steps based on similarity allows resources to be better allocated to exploring new steps, thereby improving performance while reducing computational costs.

\textbf{w/o LDB} denotes not using LDB to locate erroneous steps in our method. The average performance drop was minimal, indicating that removing LDB has little impact on our method. With execution feedback, LLMs are already capable of accurately locating errors. However, in a few cases, LDB still helps in pinpointing erroneous steps.

\subsection{Performance vs. Rollout}

We explore the results of different values of the hyperparameter rollout on Qwen3-235B-A22B, as shown in Figure \ref{fig4}. Since SRA-MCTS is prone to premature termination due to self-overestimation by the model, we set its end gate value to exceed the maximum possible score, allowing it to reach the maximum number of iterations whenever possible. we denote this variant as SRA-MCTS\_no\_eg. The results show that in the early stages, all methods exhibit significant performance improvements as the rollout increases, after which the performance gradually stabilizes. Notably, RPM-MCTS exhibits better performance even with a rollout of 1. This is because it enjoys two advantages in its first rollout: proactive guidance via its Knowledge Base during the selection phase, and wrong step truncation with rethink-based regeneration during the simulation phase. This allows it to perform at least one round of verification and reflection and generate complete code. Moreover, for simpler problems, RPM-MCTS can often arrive at the correct answer with only a single simulation, whereas traditional tree search methods tend to require multiple unnecessary expansions even for straightforward tasks.

\begin{figure}[t]
    \centering
    \includegraphics[width=0.95\columnwidth]{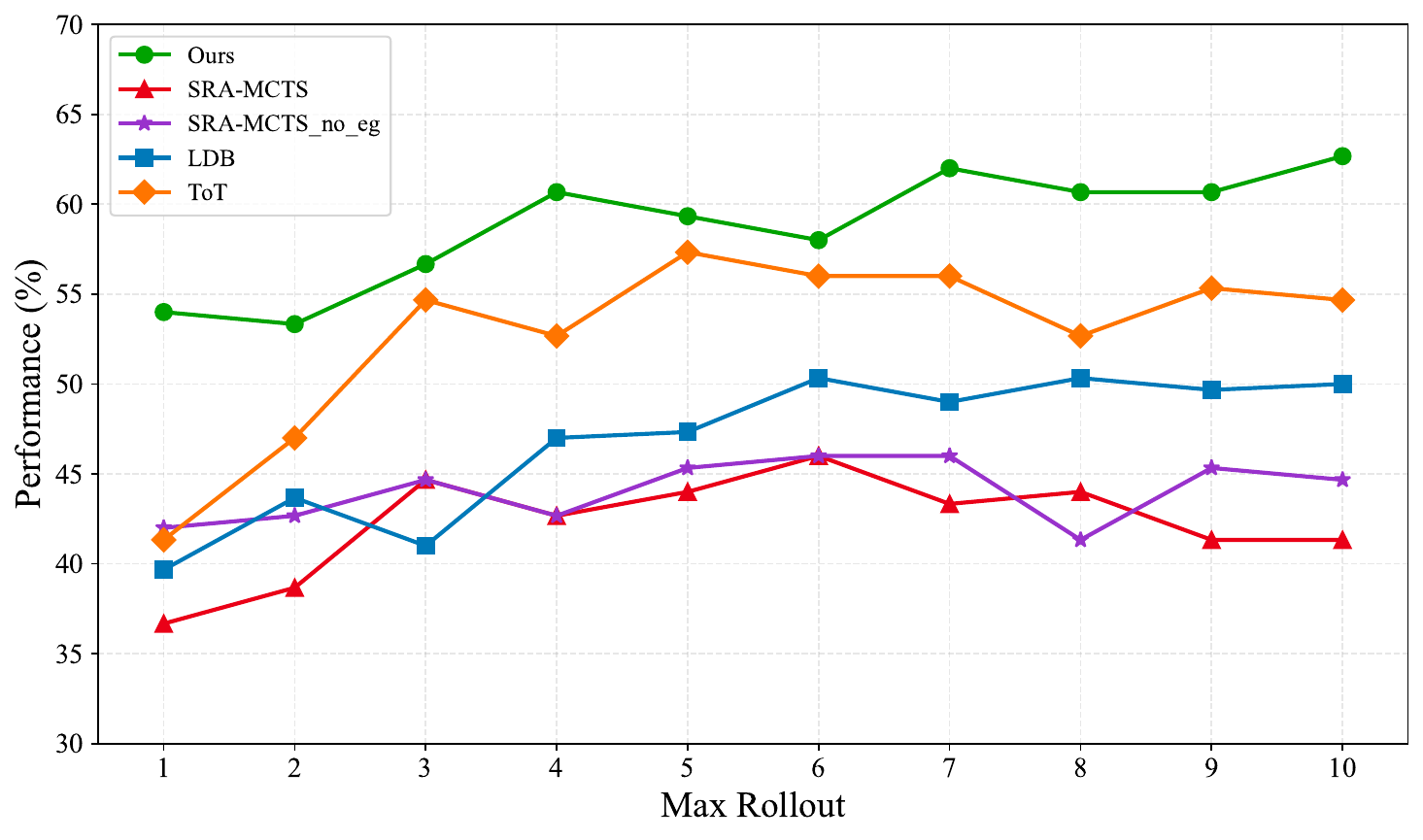} 
    \caption{Performance comparison across different maximum rollout values.}
    \label{fig4}
\end{figure}

\subsection{Token Efficiency Analysis}

Figure \ref{fig5} shows the average token usage of different methods across all benchmark datasets. Our method reduces token consumption by approximately 15\% compared to the previous MCTS method on both Qwen3-235B-A22B and Claude Sonnet 3.7. This improvement is attributed to: 1) The knowledge base retrieval scoring prioritizes more correct nodes, avoiding exploration of invalid branches. 2) Similarity filtering eliminates duplicate intermediate reasoning steps, enabling dynamic pruning of the Monte Carlo tree and reducing redundant path generation. 3) The simulation phase leverages sandbox feedback to pinpoint erroneous steps, while retaining the verified correct ones. Overall, RPM-MCTS achieves enhanced search efficiency and generation quality through knowledge base guidance and execution feedback.

\begin{figure}[t]
    \centering
    \includegraphics[width=0.95\columnwidth]{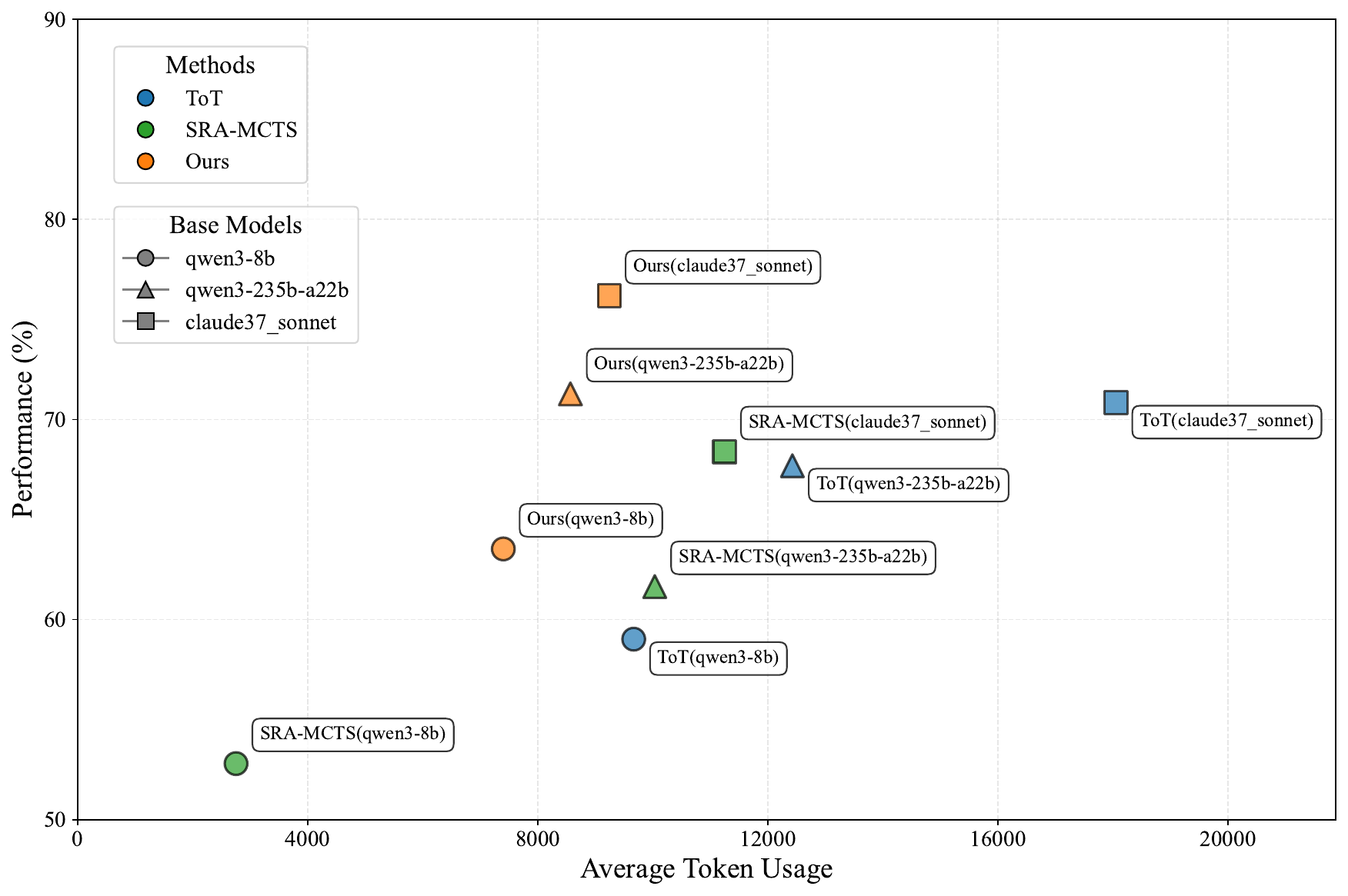} 
    \caption{Comparison of token consumption and performance across different methods and models.}
    \label{fig5}
\end{figure}

\subsection{Reasoning Steps for Data Distillation}

Our method is training-free, yet the algorithmic synthesized reasoning steps it produces can also be used for supervised fine-tuning. Based on Doubao-1.5-pro-32K model, we utilize RPM-MCTS for data distillation and construct a dataset of 2.4k code generation samples with reasoning steps. This distilled data is then combined with a foundational dataset of 170k samples to perform full fine-tuning on Doubao-1.5-pro-32K. Benchmark results in Table \ref{table2} demonstrate that the training data generated using RPM-MCTS significantly enhances the code capabilities of the base model.

\begin{table}[t]
    \centering
    \begin{tabular}{cccc}
        \hline
        \textbf{Benchmark}                                 & \textbf{Base} & \textbf{Ours} & $\Delta$ \\ \hline
        SWE-Bench (\citeauthor{jimenez2024swebench})       & 37.6          & 38.5          & +0.9     \\
        MBPP+ (\citeauthor{liu2023your})                   & 75.4          & 76.7          & +1.3     \\
        LiveCodeBench (\citeauthor{jain2024livecodebench}) & 46.2          & 50.5          & +4.3     \\
        Aider (\citeauthor{aider2024polyglot})             & 17.3          & 22.2          & +4.9     \\
        McEval (\citeauthor{mceval})                       & 57.5          & 61.2          & +3.7     \\ \hline
    \end{tabular}
    \caption{Supervised fine-tuning results on Doubao-1.5-pro-32K model using RPM-MCTS synthesized data.}
    \label{table2}
\end{table}

\section{Conclusion}
In this paper, we propose RPM-MCTS, which leverages a knowledge base and external sandbox feedback to directly obtain accurate reward values without requiring additional training of a process reward model. During the search process, errors are identified and promptly corrected. Experimental results demonstrate that RPM-MCTS outperforms current state-of-the-art methods under more constrained search budgets. Additionally, we construct training data using RPM-MCTS and perform full fine-tuning on base model, which significantly enhances code capabilities of the base model.

A limitation of RPM-MCTS is that code solvable in a single line may be divided into multiple lines due to the step-by-step approach, without impacting correctness. In the future, during the evaluation phase of MCTS, we can dynamically adjust the weights of external rewards from the knowledge base and sandbox, based on the uncertainty of LLMs.

\section{Acknowledgments}

This work was supported by the National Science and Technology Major Project (2022ZD0114803), the Natural Science Foundation of Wuhan (2023010201020229), the Fundamental Research Funds for the Central Universities (NO.NJ2023032), and the Major Program (JD) of Hubei Province (2023BAA024).

\bibliography{aaai2026}

\appendix
\onecolumn

\section{Topic Categories}

Table \ref{table1} presents the classification of algorithms in our knowledge base, which is divided into 14 categories. These categories were derived from the most common algorithmic tags on popular programming websites. The final knowledge base comprises a total of 82,923 items.

\begin{table}[h]
\centering
\begin{tabular}{@{}cc@{}}
\toprule
\textbf{Topic}                & \textbf{Nums} \\ \midrule
Data Structures               & 750           \\
Algorithm Strategies          & 1218          \\
String Processing             & 1676          \\
Sorting and Searching         & 542           \\
Graph Theory                  & 977           \\
Bit Manipulation              & 403           \\
Mathematics and Number Theory & 1658          \\
Computational Geometry        & 611           \\
Optimization Problems         & 1310          \\
Two-Pointer Techniques        & 213           \\
Dynamic Programming           & 836           \\
Recursion and Backtracking    & 226           \\
Hashing Techniques            & 316           \\
Other                         & 302           \\ \bottomrule
\end{tabular}
\caption{Knowledge base data categories and statistics.}
\label{table1}
\end{table}

\section{Dataset Details}

As detailed in Table \ref{table2}, our knowledge base was built from the CodeContests-Train and APPS-Train datasets, which together provide 11,038 training samples. The model's performance was then benchmarked against a test set consisting of six standard benchmarks: the APPS test splits (by difficulty), the CodeContests test split, HumanEval+, and MBPP+.

\begin{table}[h]
\centering
\begin{tabular}{@{}clc@{}}
\toprule
                                & \multicolumn{1}{c}{\textbf{Dataset}} & \textbf{Samples} \\ \midrule
\multirow{2}{*}{Knowledge Base} & CodeContests-Train                   & 7368             \\
                                & APPS-Train                           & 3670             \\ \cmidrule(l){2-3} 
\multirow{6}{*}{Test Set}       & APPS-Test-Introductory               & 150              \\
                                & APPS-Test-Interview                   & 150              \\
                                & APPS-Test-Competition                & 150              \\
                                & CodeContests-Test                    & 150              \\
                                & HumanEval+                           & 164              \\
                                & MBPP+                                & 378              \\ \bottomrule
\end{tabular}
\caption{Dataset statistics.}
\label{table2}
\end{table}

\section{Prompts}

Figure \ref{fig1}-\ref{fig4} demonstrate selected key prompts used in our method. The prompt in Figure \ref{fig1} generates the next step. The prompt in Figure \ref{fig2} completes full steps. The prompt in Figure \ref{fig3} analyzes execution errors to locate error steps. The prompt in Figure \ref{fig4} translates the full steps into code.

\section{Algorithm Details}

Algorithm 1 provides the detailed pseudocode for our proposed RPM-MCTS. The algorithm follows the canonical MCTS structure, iteratively performing four phases consisting of Selection, Expansion, Evaluation, and Backpropagation. A key component is the EVALUATE\_NODE function. Instead of a traditional random rollout, this function generates a complete code solution from the current path, executes it in a sandboxed environment, and assesses its correctness. If the execution fails, the algorithm activates a reflection mechanism that identifies the erroneous step, prunes the incorrect sub-path, and updates its understanding to guide future searches.

\section{Case Study}

We present a case study of our method in Figure \ref{fig5}. This figure provides a visualization of the final state of the Monte Carlo Tree after the search process concludes. In the tree, the root node represents the problem statement, while subsequent nodes correspond to individual reasoning steps. Furthermore, each node is annotated with its access sequence, visit count, and value. After four search iterations, our method successfully identifies the correct reasoning path to the solution, which corresponds to the leftmost path in the figure.

\begin{figure*}[htp]
\centering
\begin{tcolorbox}[title=Prompt for Generating Next Step]

Your task is to provide the correct \textbf{next step} based on the previous incorrect code used to solve the problem and a reflection, for a given programming problem and its existing solution steps (which are incomplete).
Let's think step by step. But you only generate one step at a time.
We aim to decompose complex problems into a series of simpler subproblems and sequentially generate the corresponding steps to solve each subproblem.
All the substeps should be combined in a way that avoids contradictions, forming a coherent solution to the original complex problem.

\hrulefill

\textbf{Input format (n steps):}
\begin{itemize}
    \item \textcolor{my_purple}{\textbf{Problem:}} \{problem\}
    \item \textcolor{my_purple}{\textbf{Existing steps:}} \{existing\_steps\}
    \item \textcolor{my_purple}{\textbf{Analysis:}} \{reflection\}
    \item \textcolor{my_purple}{\textbf{History:}} \{history\}
\end{itemize}
The historical content is the solution proposed earlier. To ensure the diversity of solutions, please do not generate ideas identical to those in the historical content.

\hrulefill

\textbf{Guidelines:}
\begin{itemize}
    \item The steps you generate will be passed to a code generation model, so they should be structured in a way that is easy for the model to understand.
    \item Keep each step concise and focused, avoiding the inclusion of too much information at once. Ensure clear organization and logical progression in your reasoning.
    \item \textbf{Important:} You can use very little code as detailed explanations in your answers, but you \textbf{cannot} just write code.
    \item If your answer includes code, it will cause unforeseen losses!
    \item Your answer should be based on the given analysis. Only if the analysis is wrong can you answer it in your own way.
    \item If no existing steps are provided, you should output the first step based on the given analysis.
    \item If there are existing steps, output the next step (Step n+1) that logically follows the provided analysis and the previous steps.
\end{itemize}

\hrulefill

\textbf{Output format:}
\begin{itemize}
    \item \textcolor{my_green}{\textbf{Next step:}} ...
\end{itemize}
Your response should only generate solutions to the problem, without any extra words.

\end{tcolorbox}
\caption{Prompt for Generating Next Step.}
\label{fig1}
\end{figure*}

\begin{figure*}[htp]
\centering
\begin{tcolorbox}[title=Prompt for Generating Full Steps]

Your task is to take a programming problem and incomplete solution steps (not a full answer), then continue from the provided steps to complete all remaining steps and generate the complete final solution.

Let's think step by step. We aim to decompose complex problems into a series of simpler subproblems and sequentially generate the corresponding steps to solve each subproblem.
All the substeps should be combined in a way that avoids contradictions, forming a coherent solution to the original complex problem.
Note: Do not modify the existing solution steps.

\hrulefill

\textbf{Input format (n steps):}
\begin{itemize}
    \item \textcolor{my_purple}{\textbf{Problem:}} \{problem\}
    \item \textcolor{my_purple}{\textbf{Existing steps:}} \{existing\_steps\}
\end{itemize}

\hrulefill

\textbf{Guidelines:}
\begin{itemize}
    \item If n is equal to 0, you need to start from scratch and analyze the solution idea briefly, and then output the complete answer.
    \item Otherwise, you need to output the complete answer that you think is correct following the train of thought of the existing steps.
    \item Each step generated should be concise and focused, addressing only a small part of the solution. Avoid making the steps too complex or combining multiple ideas into one.
    \item The complete solution should consist of at least three steps, so don't skip any essential steps.
    \item Your output should be clear and systematic, with each step described one at a time to ensure logical progression.
    \item Note: You are only allowed to describe the reasoning steps in natural language. Do not output any code.
\end{itemize}

\hrulefill

\textbf{Output format:}
\begin{itemize}
    \item \textcolor{my_green}{\textbf{Step 1:}} ...
    \item \textcolor{my_green}{\textbf{Step 2:}} ...
    \item ...
    \item \textcolor{my_green}{\textbf{Step n:}} ...
    \item \textcolor{my_green}{\textbf{Step n + 1:}} ...
    \item ...
\end{itemize}
Among them, Step 1 to Step n are consistent with the existing steps. Continue to generate based on the existing steps to obtain a complete answer.
The following is the input. Please output according to the specified output format, do not output unnecessary information, and do not repeat the question.
Note: Your output should start from Step 1 and include all the steps, not just the next step.

\end{tcolorbox}
\caption{Prompt for Generating Full Steps.}
\label{fig2}
\end{figure*}

\begin{figure*}[htp]
\centering
\begin{tcolorbox}[title=Prompt for Code Debugging and Analysis]

The following is a Python code problem, which includes the thoughts and code for solving the problem, as well as the return results of debugging for a failed test case.

\hrulefill

\textbf{Input:}
\begin{itemize}
    \item \textcolor{my_purple}{\textbf{Python code problem:}} \{problem\}
    \item \textcolor{my_purple}{\textbf{Thoughts:}} \{solution\}
    \item \textcolor{my_purple}{\textbf{Code:}} \{code\}
    \item \textcolor{my_purple}{\textbf{Test case debug information:}} \{exec\_result\}
\end{itemize}
The debugging process is to first split the code into block-level code according to the AST. If the block-level code is correct after debugging analysis, the "correct" field is True, otherwise it is False. The "explanation" field is the analysis of the block-level code debugging.

\hrulefill

\textbf{Guidelines:}
Your task is to determine which specific step is written incorrectly based on the debug return results and conduct an analysis and summary. The correctly generated code and corresponding thought processes will be retained, while the incorrect code and corresponding thought processes will be discarded. You need to analyze and summarize the points to note so that subsequent thought processes can be generated based on the correct thought processes to correct the previous errors.

\hrulefill

\textbf{Output format:}
Your output consists of two parts:
\begin{itemize}
    \item \textbf{1.} Which specific step went wrong. Wrap it with the \texttt{<step\_n>x</step\_n>} XML tag, where x represents the specific number of the first erroneous step. If there are multiple erroneous steps in the thought process, only output the number of the first erroneous step. Do not output any extra content.
    \item \textbf{2.} Analyze and summarize the points to note.
\end{itemize}
The final output should look like this: \textcolor{my_green}{\texttt{<step\_n>x</step\_n>...}}, where ... represents the generated analysis.

\end{tcolorbox}
\caption{Prompt for analyzing debugging results and identifying errors.}
\label{fig3}
\end{figure*}

\begin{figure*}[htp]
\centering
\begin{tcolorbox}[title=Prompt for Code Implementation]

You will play the role of a code implementer, writing a complete code based on the given problem and the step-by-step analysis of the problem.
Your code must strictly follow the analysis steps provided and should not include your own opinions.

\hrulefill

\textbf{Rules:}
\begin{itemize}
    \item Importing function libraries(like: import math) and output function code only, without main function so that I can call your generated functions directly.
    \item The output code should be wrapped with code blocks (like ```python). Example: ```python\textbackslash{}ndef add(a, b):\textbackslash{}n    return a + b\textbackslash{}n```.
\end{itemize}

\hrulefill

\textbf{Input:}
\begin{itemize}
    \item \textcolor{my_purple}{\textbf{question:}} \{question\}
    \item \textcolor{my_purple}{\textbf{analysis:}} \{analysis\}
\end{itemize}

\end{tcolorbox}
\caption{Prompt for generating code based on steps.}
\label{fig4}
\end{figure*}

\begin{algorithm*}[ht]
\caption{The RPM-MCTS Algorithm for Code Generation}
\label{alg:rpm-mcts}
\textbf{Input}: Problem description $P$, total iterations $I$, branching factor $B$, success threshold $\theta_{succ}$\\
\textbf{Output}: The best generated code solution $C_{best}$
\begin{algorithmic}[1] 
\STATE $v_{root} \gets \text{CREATE\_NODE}(P)$
\FOR{$i \gets 1$ to $I$}
    \STATE \texttt{// 1. Selection}
    \STATE $v_l \gets v_{root}$
    \WHILE{$v_l$ is fully expanded}
        \STATE $v_l \gets \text{SELECT\_BEST\_CHILD\_UCB}(v_l)$
    \ENDWHILE
    \STATE
    \STATE \texttt{// 2. Expansion}
    \IF{$v_l$ is not a terminal node}
        \STATE $S_{gen} \gets \emptyset, H_{hist} \gets \emptyset$
        \FOR{$j \gets 1$ to $B$}
            \STATE $s_{new} \gets \text{GENERATE\_NEXT\_STEP}(P, \text{path}(v_l), H_{hist})$ \COMMENT{Generate diverse steps}
            \STATE Add $s_{new}$ to $S_{gen}$ and $H_{hist}$
        \ENDFOR
        \STATE $S_{unique} \gets \text{FILTER\_SIMILAR\_STEPS}(S_{gen})$ \COMMENT{Filter semantic duplicates}
        \FOR{each step $s$ in $S_{unique}$}
            \STATE $v_c \gets \text{ADD\_CHILD}(v_l, s)$
            \STATE $v_c.value \gets \text{GET\_INITIAL\_VALUE}(\text{path}(v_c))$ \COMMENT{Score from knowledge base \& LLM}
        \ENDFOR
        \STATE Mark $v_l$ as fully expanded
    \ENDIF
    \STATE
    \STATE \texttt{// 3. Evaluation (Simulation)}
    \STATE $v_r \gets \text{SELECT\_BEST\_CHILD\_UCB}(v_l)$
    \IF{$v_r$ is not NULL}
        \STATE $is\_solved, Q_{final} \gets \text{EVALUATE\_NODE}(v_r, P, \theta_{succ})$
        \IF{$is\_solved$}
            \STATE \textbf{return} $\text{GET\_SOLUTION}(v_r)$ \COMMENT{Optimal solution found, terminate early}
        \ENDIF
    \ELSE
        \STATE $Q_{final} \gets v_l.value$ \COMMENT{Use parent value if no children to evaluate}
    \ENDIF
    \STATE
    \STATE \texttt{// 4. Backpropagation}
    \STATE $\text{BACKPROPAGATE}(v_r, Q_{final})$
\ENDFOR
\STATE \textbf{return} $\text{GET\_BEST\_SOLUTION}(v_{root})$ \COMMENT{Return best solution after all iterations}
\STATE
\STATE \textbf{Function} EVALUATE\_NODE($v, P, \theta_{succ}$)
    \STATE $\pi_s \gets \text{path}(v)$
    \STATE $S_{full}, C \gets \text{GENERATE\_FULL\_SOLUTION}(\pi_s)$
    \STATE $r_{exec}, res_{sb} \gets \text{EXECUTE\_CODE}(C)$ \COMMENT{Evaluate in sandbox}
    \STATE $r_{llm}, f \gets \text{EVALUATE\_WITH\_LLM}(S_{full}, C, res_{sb})$ \COMMENT{$f$ is reflection}
    \STATE $Q_{comb} \gets \gamma \cdot r_{exec} + (1-\gamma) \cdot r_{llm}$ \COMMENT{Weighted combined value}
    \IF{$r_{exec}$ is SUCCESS and $r_{llm} \ge \theta_{succ}$}
        \STATE $\text{ADD\_SOLUTION\_TO\_TREE}(v, S_{full})$
        \STATE \textbf{return} true, $Q_{comb}$
    \ENDIF
    \IF{$r_{exec}$ is FAILURE}
        \STATE $idx_{err} \gets \text{LOCATE\_ERROR\_STEP}(S_{full}, res_{sb}, f)$
        \STATE $S_{pruned} \gets \text{PRUNE\_STEPS}(S_{full}, idx_{err})$
        \STATE $\text{ADD\_SOLUTION\_TO\_TREE}(v, S_{pruned})$ \COMMENT{Add correct partial path}
        \STATE $\text{UPDATE\_REFLECTION\_IN\_TREE}(v, f)$
    \ENDIF
    \STATE \textbf{return} false, $Q_{comb}$ \COMMENT{Return failure if not solved}
\end{algorithmic}
\end{algorithm*}

\begin{figure*}[ht]
\centering
\includegraphics[width=0.7\textwidth]{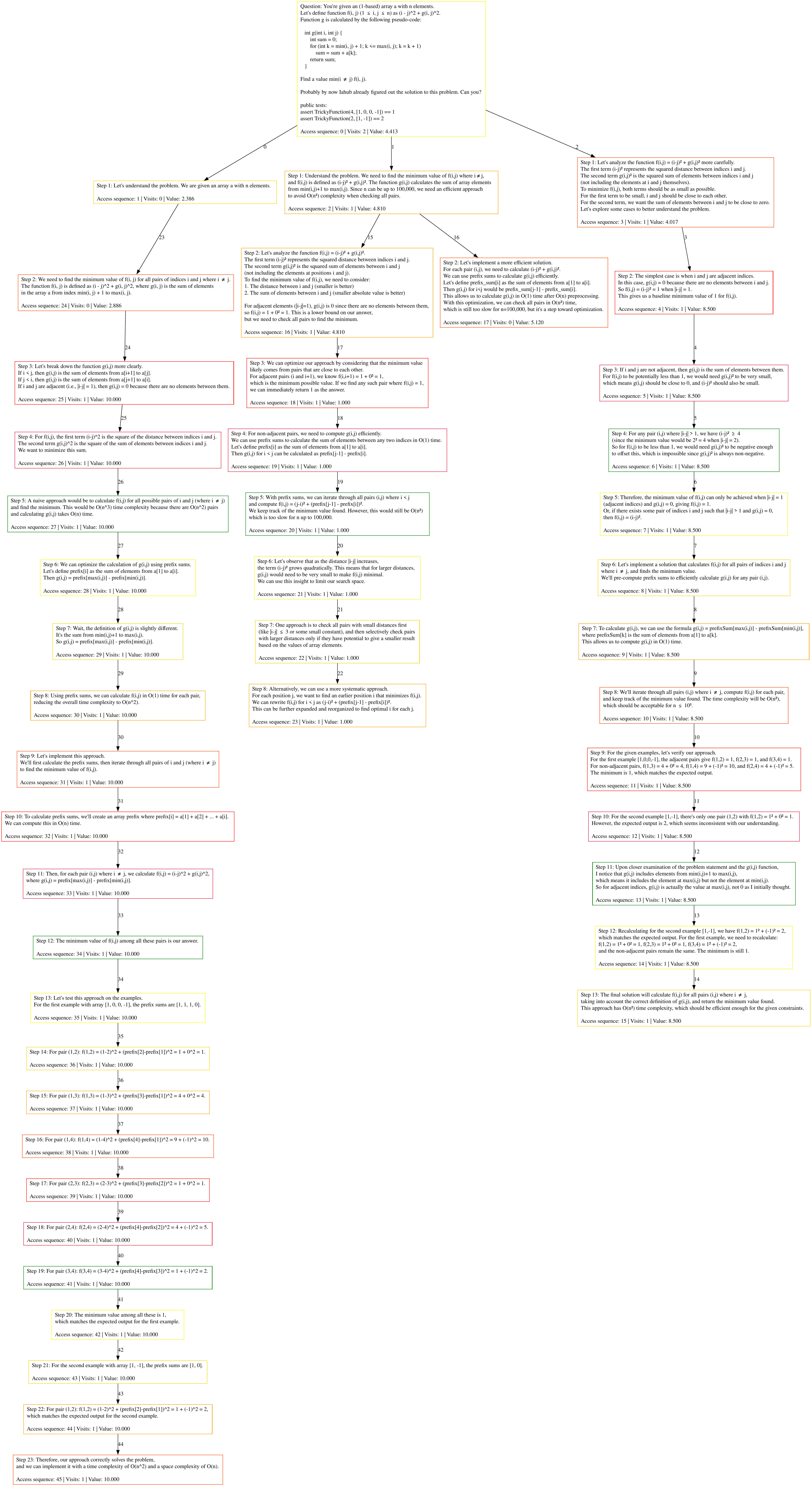} 
\caption{Case study.}
\label{fig5}
\end{figure*}

\end{document}